\documentclass[11pt]{article}

\usepackage[final]{acl}

\usepackage{times}
\usepackage{latexsym}
\usepackage{hyperref}      

\usepackage[T1]{fontenc}

\usepackage[utf8]{inputenc}

\usepackage{microtype}

\usepackage{inconsolata}

\usepackage{graphicx}
\usepackage{amsmath}

\title{MahaParaphrase: A Marathi Paraphrase Detection Corpus \\ and BERT-based Models}

\author{
\textbf{Suramya Jadhav}\textsuperscript{1,4}, \textbf{Abhay Shanbhag}\textsuperscript{1,4}, 
\textbf{Amogh Thakurdesai}\textsuperscript{1,4}, \\ \textbf{Ridhima Sinare}\textsuperscript{1,4},
\textbf{Ananya Joshi}\textsuperscript{2,4}, and \textbf{Raviraj Joshi}\thanks{Correspondence: ravirajoshi@gmail.com}\textsuperscript{3,4} \\
\textsuperscript{1}Pune Institute of Computer Technology, Pune \\
\textsuperscript{2}MKSSS’ Cummins College of Engineering for Women, Pune \\
\textsuperscript{3}Indian Institute of Technology Madras, Chennai \\
\textsuperscript{4}L3Cube Labs, Pune \\
}
\begin{document}
\maketitle
\begin{abstract}
Paraphrases are a vital tool to assist language understanding tasks such as question answering, style transfer, semantic parsing, and data augmentation tasks. Indic languages are complex in natural language processing (NLP) due to their rich morphological and syntactic variations, diverse scripts, and limited availability of annotated data. In this work, we present the L3Cube-MahaParaphrase Dataset, a high-quality paraphrase corpus for Marathi, a low resource Indic language, consisting of 8,000 sentence pairs, each annotated by human experts as either Paraphrase (P) or Non-paraphrase (NP). We also present the results of standard transformer-based BERT models on these datasets. 
The dataset and model are publicly shared at 
\href{https://github.com/l3cube-pune/MarathiNLP}{https://github.com/l3cube-pune/MarathiNLP}.

\end{abstract}

\section{Introduction}
Paraphrasing is the task of generating semantically equivalent sentences with different wording or structure. \citep{bhagat2013paraphrase} defines paraphrases as "different surface realizations of the same semantic content, while \citet{barzilay2001extracting} describes paraphrases as "textual expressions that share the same meaning but differ in form". It plays a crucial role in various natural language processing (NLP) applications and is inherently familiar to speakers of all languages \citep{madnani2010generating}. 
Paraphrasing can be a vital tool to assist language understanding tasks such as question answering, style transfer  \citep{krishna-etal-2020-reformulating}, semantic parsing  \citep{cao-etal-2020-unsupervised-dual}, and data augmentation tasks \citep{gao-etal-2020-paraphrase}. Interestingly, paraphrase identification can also be effectively implemented for plagiarism detection \citep{hunt2019machine}.

The MRPC \footnote{\href{https://www.microsoft.com/en-us/download/details.aspx?id=52398}{Microsoft Research Paraphrase Corpus}}, an English paraphrase corpus, is one such dataset that set a benchmark in creating paraphrase datasets. Since then, a wide variety of techniques, as mentioned in \citet{zhou-bhat-2021-paraphrase}, \citet{madnani2010generating}, \citet{gadag2016review} have been developed. However, many of such developments have been around the English language, which for a long time now has been a high-resource language. With plenty of corpora spanning multiple domains like news, sentiment analysis, etc., the preliminary source of sentences becomes rich in diversity, making paraphrase data generation easier. Moreover, models used for detecting semantic and lexical relations between sentence pairs are extensively being developed and put into use, as in \citet{khairova2022using}. Developments have also been around languages like Vietnamese \citep{phan2022vietnamese} and Finnish (The Turku Paraphrase \citep{kanerva2024towards}). The ParaCotta corpus \citep{aji2022paracottasyntheticmultilingualparaphrase} consists of a paraphrase dataset for around 17 languages and also illustrates how Sentence Transformers like S-BERT can be effectively used for generation as well as evaluation.

While the most important thing to build any model or task-specific dataset (a paraphrase dataset in this case) is having a diverse corpus of scraped and manually verified data, this is severely lacking in the case of Indic languages.
This is because of the complexity of Indic languages due to their rich morphological and syntactic variations, diverse scripts, and limited availability of annotated data. However, there has been significant progress in Indic NLP research due to the AI4Bharat-IndicNLP  project and IndicNLPSuite \citep{kakwani2020indicnlpsuite}, who provide corpora and resources like pretrained models for 10 Indian languages across tasks like sentiment analysis and news headline classification. The Amritha corpus is a paraphrase dataset focused on 4 languages: Hindi, Malayalam, Punjabi, and Tamil \citep{anand2016shared}. The BanglaParaphrase \citep{akil2022banglaparaphrase} focuses on using IndicBART for curating the Bangla paraphrase corpus.

Very few research groups such as L3Cube \footnote{\href{https://github.com/l3cube-pune/MarathiNLP}{https://github.com/l3cube-pune/MarathiNLP}} are focusing on regional, low resource Indic languages like Marathi. They have also demonstrated that using LLMs for dataset curation (for annotations) has not shown promising results \citep{jadhav2025limitationsllmannotatorlow}. Moreover, handling paraphrases in Marathi is tricky due to its lexical syntax, complex linguistic features, and the influence of various dialects \citep{lahoti2022survey}, \citep{dani2024review}. L3Cube's MahaNLP project \citep{joshi2022l3cube}, focused specifically on the Marathi language by developing a Marathi corpus across multiple domains, which helps Marathi NLP.

Contributing to the same project, in this work, we present the L3Cube-MahaParaphrase\footnote{\href{https://huggingface.co/datasets/l3cube-pune/MahaParaphrase}{https://huggingface.co/datasets/l3cube-pune/MahaParaphrase}} Dataset, a robust Marathi paraphrase corpus with each sentence pair annotated and manually curated as Paraphrase (P) or Nonparaphrase (NP) with a total of 8K sentence pairs. We further divide the dataset into 5 buckets based on the increasing degree of paraphrase with word overlap and semantic accuracy as factors, giving future research a chance to explore based on varying degrees of paraphrase. The 2-label annotation approach employed is thoroughly described. Furthermore, we also present the results of standard transformer-based BERT models on these datasets.
Our key contributions are as follows:
\begin{itemize}
    \item Created a gold standard 8K Paraphrase corpus for Marathi with labelled sentences pairs as P or NP (4K each for P and NP).
    \item We divide the MahaParaphrase corpus into multiple buckets based on lexical (word-level) overlap and semantic similarity, thereby capturing varying degrees of paraphrastic and non-paraphrastic relationships between sentence pairs.
    \item We evaluate existing models like Muril, mBERT, IndicBERT as well as L3Cube's MahaBERT for benchmarking. Additionally, we release MahaParaphrase-BERT\footnote{\href{https://huggingface.co/l3cube-pune/marathi-paraphrase-detection-bert}{https://huggingface.co/l3cube-pune/marathi-paraphrase-detection-bert}}, a fine-tuned version of MahaBERT trained on the MahaParaphrase corpus.
\end{itemize}

\section{Literature Review}
Research on paraphrase detection and generation has been extensively explored in high-resource languages like English. Different techniques have emerged for paraphrase generation and detection, ranging from using Bi-LSTM with pretrained GLoVe word vectors \citep{shahmohammadi2021paraphrase} to fine-tuning T5 models \citep{kubal2021unified, palivela2021optimization}, and using advanced transformer models like GPT and BERT \citep{natsir2023deep}. Combined techniques like variational sampling with hashing sampling, an unsupervised method, have been used for phrase-level and sentence-level paraphrase detection \citep{hejazizo2021combining}. \citet{gangadharan2020paraphrase} demonstrated how word vectorization can convert textual data into numerical representations for paraphrase detection and analysis, exploring Count Vectorizer, Hashing Vectorizer, TF-IDF Vectorizer, FastText, ELMo, GloVe, and BERT.

It is important to note that much of this research primarily used English paraphrase corpora for experimentation. Experimentation for other languages is limited due to the lack of quality datasets.

As far as low-resource paraphrase datasets are concerned, \citet{kanerva2024towards} introduced a comprehensive dataset, 'Turku Paraphrase,' for the Finnish language. The \href{https://huggingface.co/datasets/GEM/opusparcus}{OpenParcus Dataset} consists of paraphrases for six European languages. The ParaCotta Corpus \citep{aji2022paracottasyntheticmultilingualparaphrase}, which includes around 17 languages, including Hindi, is one of the most diverse datasets spanning a wide variety of languages.

Talking about Indic languages, generation of paraphrases becomes difficult because of rich morphological and syntactic variations and diverse scripts. Moreover, all Indic languages fall under the low-resource category due to the lack of annotated data. The Bangla Paraphrase \citep{akil2022banglaparaphrase} uses IndicBART to synthetically generate paraphrases. In \citet{anand2016shared}, a significant milestone was achieved with the release of the Amritha paraphrase corpus for four Indic languages: Hindi, Malayalam, Punjabi, and Tamil, as part of the DPIL@FIRE2016 Shared Task, enabling participants to experiment further. 

Another notable effort is the IndicParaphrase Dataset by AI4Bharat\citep{Kumar2022IndicNLGSM}, which includes 11 Indic languages: Assamese (as), Bengali (bn), Gujarati (gu), Kannada (kn), Hindi (hi), Malayalam (ml), Marathi (mr), Oriya (or), Punjabi (pa), Tamil (ta), and Telugu (te).
This dataset provides input and target sentences, as well as a reference list of five sentences with different levels of lexical correlation.

While the Marathi subset in IndicParaphrase is huge for Marathi, it is important to note that it consists only of paraphrased sentence pairs. The same applies to many of the above-mentioned corpora, like BanglaParaphrase, Turku Paraphrase, and OpusParcus. However, the Amritha corpus for the DPIL@FIRE2016 shared task includes labeled sentence pairs as P (Paraphrase) and NP (Non-paraphrase) but does not include Marathi. To date, there is no Marathi paraphrase dataset that consists of both P and NP sentence pairs with 5 varying paraphrastic levels.

\begin{figure*}
    \centering
    \includegraphics[width=1\linewidth]{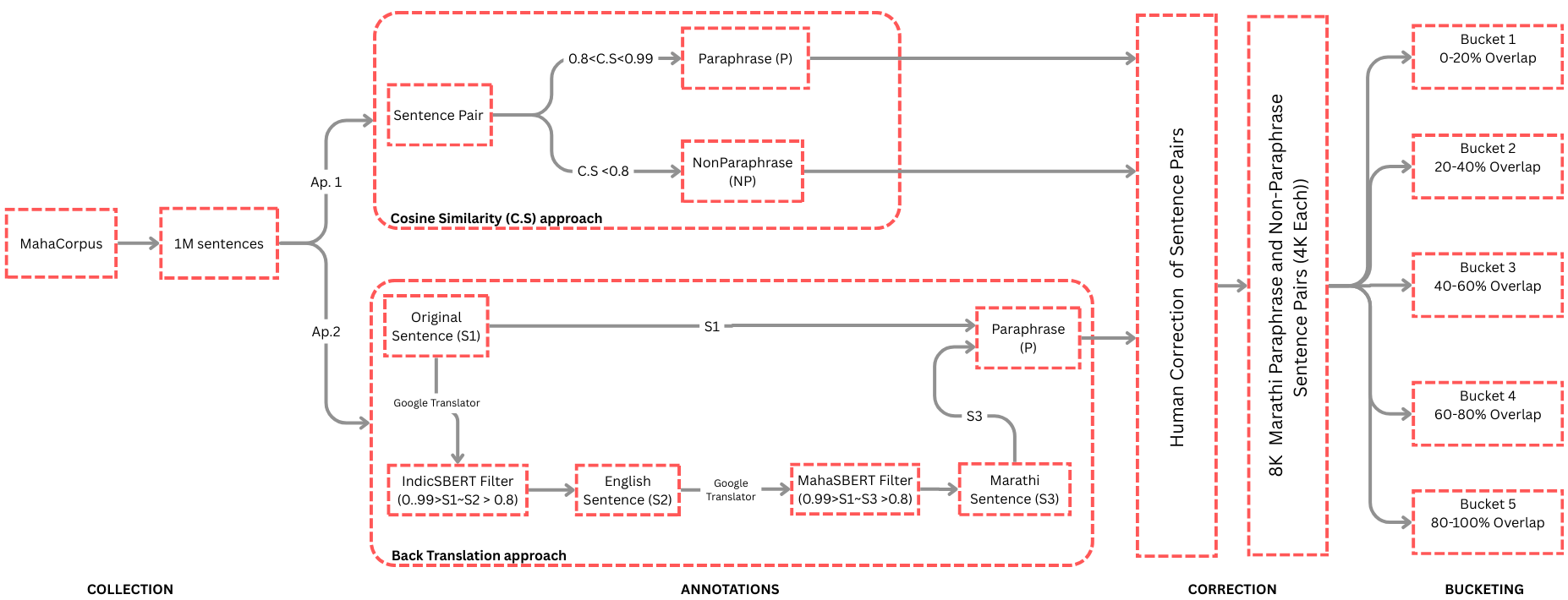}
    \caption{MahaParaphrase Dataset Curation Workflow.}
    \label{fig:Dataset_Workflow}
\end{figure*}
\section{Dataset}
This section provides information on how the dataset was collected. We created the paraphrase dataset in three phases: gathering sentences from MahaCorpus, categorizing them into P and NP using both cosine similarity and back-translation approaches, and then these sentences were manually verified for errors by four native Marathi human annotators. Finally, the sentences were divided into five equally distributed buckets based on word overlaps. Each of these steps is discussed in the following subsections and represented in Figure \ref{fig:Dataset_Workflow}.

\subsection{Collection}
The required Marathi sentences were taken from the MahaCorpus dataset by L3 Cube, which spans a wide range of topics, including news, sentiment, and hate speech. These sentences were collected from various news sources from the Maharashtra region.

We randomly selected 1 million sentences from this corpus as our primary dataset. In this section, we elaborate on the annotation process for labeling sentences as Paraphrase (P) and Non-Paraphrase(NP).
\subsection{Annotations}
The collected sentence pairs were annotated using 2 approaches so as to get a mixture of both real and synthetic data. We now explain the two approaches used to categorize sentences as P or NP.

\subsubsection{Approach 1: Cosine Similarity
}
In this approach, we calculated the cosine similarity for every pair of sentences from the 1 million collected sentences.
Since contextualized token embeddings have been shown to be effective for paraphrase detection \citep{devlin2019bert}, we use BERTScore \citep{zhang2020bertscoreevaluatingtextgeneration} to ensure semantic similarity between the source and candidates.  
To do this, we used the sentence transformer MahaSBERT \citep{joshi2022l3cubemahasberthindsbertsentencebert} to generate sentence embeddings. Then, we calculated the cosine similarity scores between the embeddings of each pair of sentences.

Based on the scores, we categorized the sentences as follows:

\begin{itemize}
    \item  If the cosine similarity (C.S) score was less than 0.8, the sentences were labeled as NP.
    \item  If the cosine similarity score was between 0.8 and 0.99, the sentences were labeled as P.
\end{itemize}

\subsubsection{Approach 2: Back Translation}
In this approach, we used the back-translation technique to generate paraphrase sentences. The process involves:
\begin{itemize}
     
    \item Translating a Marathi sentence (S1) into English (S2) using Google Translator.
    \item Translating it back from English (S2) to Marathi (S3) using Google Translator.
\end{itemize}

This gives us a pair of sentences: the original sentence (S1) and the back-translated sentence (S3), which we consider as paraphrases.
\\ 

\textbf{Filter:} To ensure that S3 is not identical to S1, we applied a filter after translation.
We used a sentence transformer to calculate the cosine similarity between the sentences and enforced the following rules:

\begin{itemize}
    \item If the cosine similarity (C.S) score was less than 0.8, we discarded the pair (indicating the meaning might have changed).
    \item If the cosine similarity score was greater than 0.99, we discarded the pair (indicating the sentences were too similar, likely identical, and not valid paraphrases).
\end{itemize}

For Marathi-to-English translations (i.e. S1 to S2), we used IndicSBERT \citep{deode-etal-2023-l3cube}, and for Marathi-to-Marathi comparisons, we used MahaSBERT \citep{joshi2022l3cubemahasberthindsbertsentencebert} to compute the cosine similarity score between the two sentences (i.e S1 and S3) in the filter.
\\ 
\\
\textbf{Combining Approaches:}  Approach 1 provides real data, as both sentences are directly taken from the MahaCorpus \citep{joshi2022l3cube1}. On the other hand, Approach 2 generates new sentences, which are synthetic. To maintain balance, we used an equal number of sentences from both approaches.
\subsection{Human Correction}
To ensure that all sentences were correctly classified, the entire dataset was manually verified by native Marathi speakers proficient in reading and writing Marathi.

Any errors were corrected by manually modifying the sentences to ensure accuracy and consistency.
\subsection{Bucketing}

We further categorized the P and NP data into five buckets for each category, based on word overlap. 
\begin{itemize}
    \item Bucket B5: 80-100\% word overlap
    \item Bucket B4: 60-80\% word overlap
    \item Bucket B3: 40-60\% word overlap
    \item Bucket B2: 20-40\% word overlap
    \item Bucket B1: 0-20\% word overlap
\end{itemize}

\noindent Word overlap is calculated as:
\begin{equation}
\text{Word Overlap}(A, B) = \frac{|W(A) \cap W(B)|}{|W(A) \cup W(B)|}
\end{equation}
where $W(A)$ and $W(B)$ denote the sets of words in sentences $A$ and $B$ respectively.

For example, a pair of sentences in B5 of the NP dataset will be semantically different (and hence categorized as NP), even though they have around 80\% or more word overlap. This is significant because it highlights that even with high word overlap, sentences can have different meanings, emphasizing the importance of word order when considering paraphrase pairs.

Now consider another example: a pair of sentences in B1 of the P dataset. These sentences have low word overlap but are still considered paraphrases. This shows that sentences with different words (such as synonyms) can also form valid paraphrase pairs.

This categorization into buckets makes the dataset robust and versatile for evaluation across different scenarios, such as high-overlap non-paraphrase sentence pairs or low-overlap paraphrase sentence pairs. Refer table \ref{tab:dataset_table} for bucket-wise examples from the dataset.

\section{Dataset Statistics}

\begin{table*}[htbp]
  \centering
  \includegraphics[page=1,width=1\textwidth]{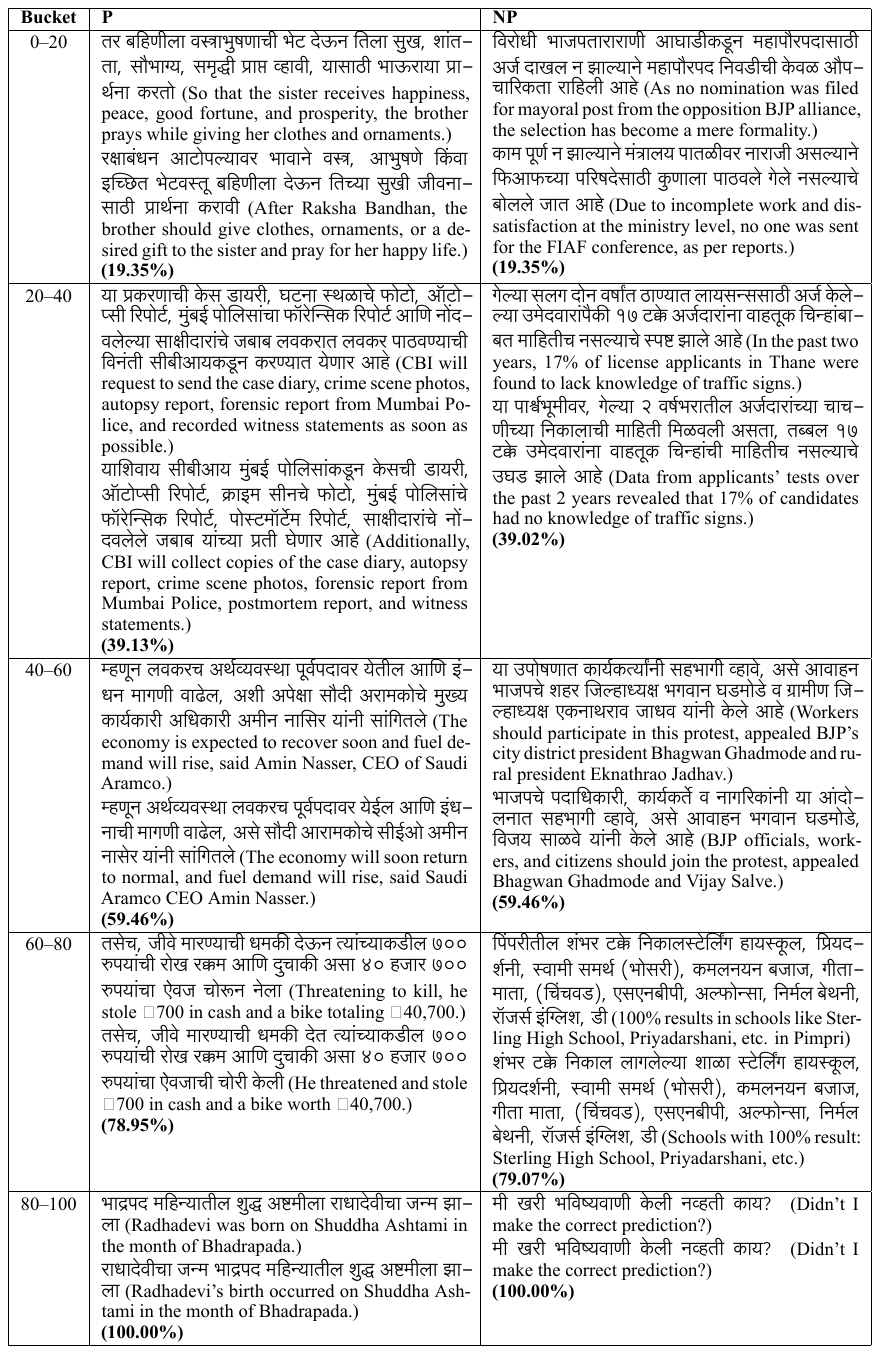}
  \caption{P and NP Sentence Pairs with Overlap Percentages by Bucket. Every cell consists of S1 and S2 along with their english translations followed by the word overlap percentages (in \textbf{bold}). The examples choosen are pairs with max word overlap in that particular bucket.}
  \label{tab:dataset_table}
\end{table*}
The dataset contains 4000 rows of sentence pairs labeled as paraphrase (P), and 4000 rows of sentence pairs labeled as nonparaphrase (NP). Each row in the dataset contains a sentence along with a paraphrase or non-paraphrase label. 

The average word count for sentences in the dataset, as well as the difference between the averages, is given in Table \ref{tab:avg_word_counts}.

\begin{figure}[ht]
    \centering
    \includegraphics[width=0.99\linewidth]{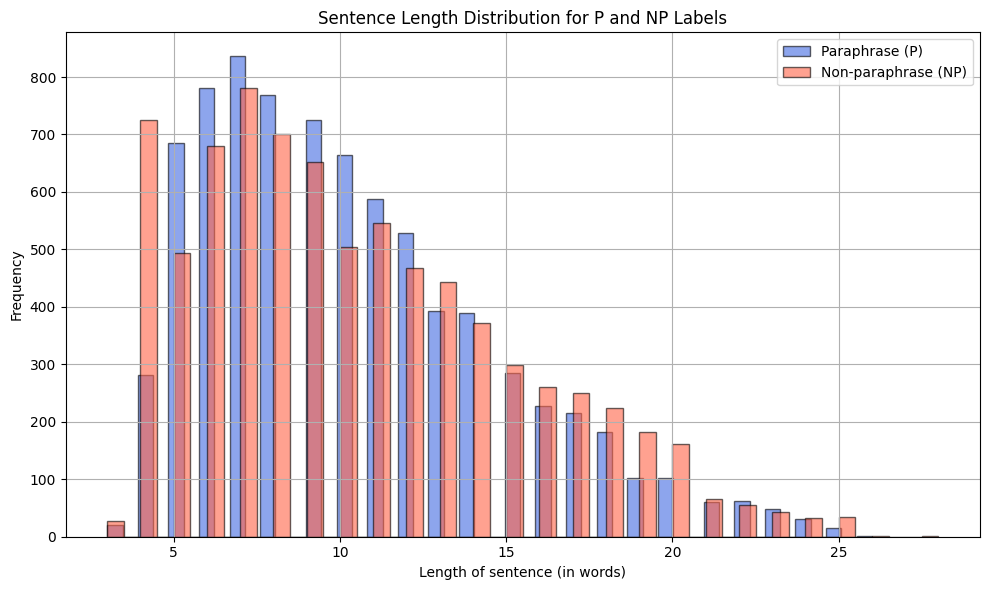}
    \caption{Sentence length distribution for Paraphrase (P) and Non-paraphrase (NP) classes. The x-axis shows sentence length in words, and the y-axis indicates the frequency of those lengths.}
    \label{fig:sentence_length_distribution}
\end{figure}
\begin{figure}[ht]
    \centering
    \includegraphics[width=0.99\linewidth]{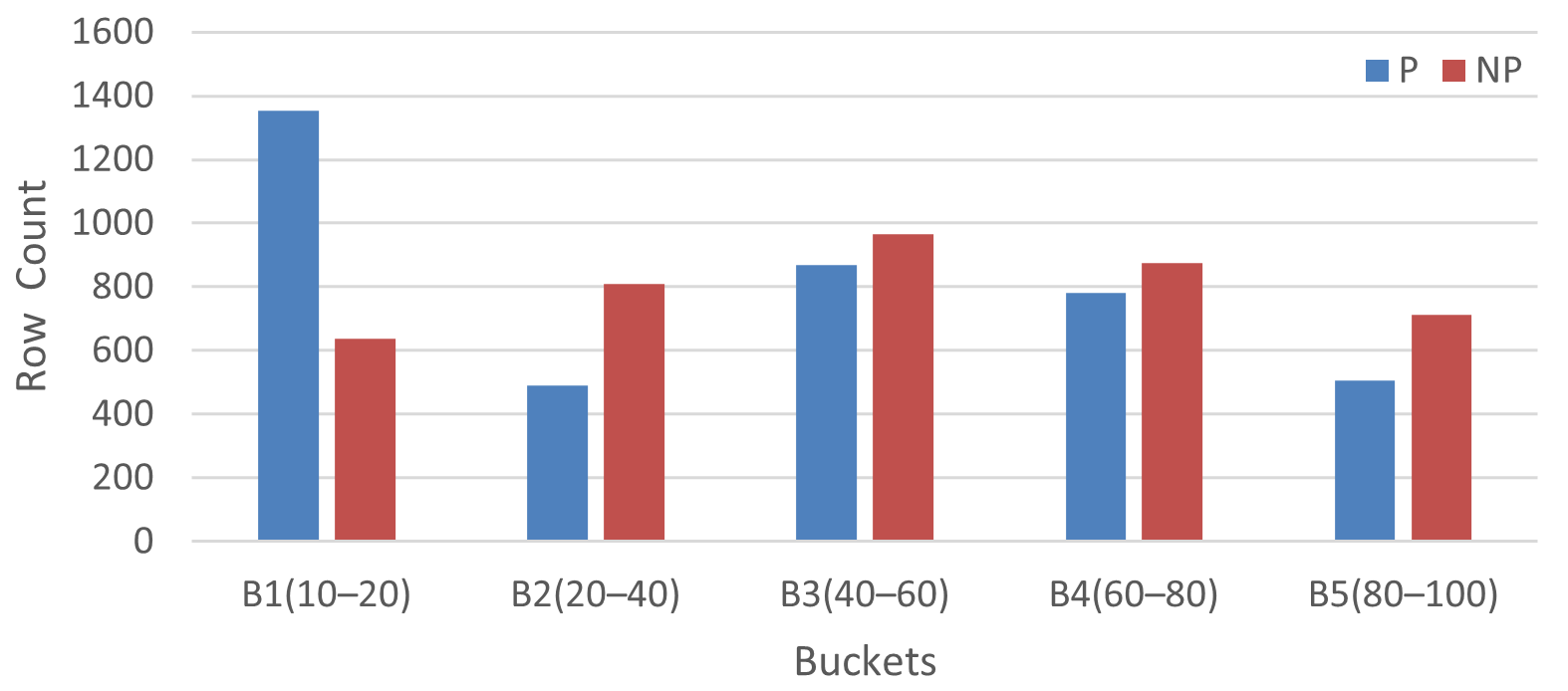}
    \caption{Bucket Wise Distribution. The values in brackets are the word overlap percentages for each bucket. }
    \label{fig:bucket_row_count}
\end{figure}

Figure \ref{fig:sentence_length_distribution} shows the distribution of sentence lengths for both Paraphrase (P) and Non-paraphrase (NP) classes in our dataset. Both distributions follow a similar pattern, with the highest frequency occurring for sentence lengths between 5 and 15 words.

Figure \ref{fig:bucket_row_count} shows the bucket wise row count distribution for both Paraphrase (P) and Non-paraphrase (NP) classes. While the total count of P and NP are same (i.e 4000 each),  their distribution across buckets is varied.

\begin{table}[ht]
\centering
\resizebox{\columnwidth}{!}{%
\begin{tabular}{lccc}
\hline
\textbf{Dataset} & \textbf{Sentence 1 Avg.} & \textbf{Sentence 2 Avg.} & \textbf{Avg. Diff.} \\
\hline
Paraphrase  & 10.43 & 9.99 & 1.45 \\
Non-Paraphrase & 9.52  & 11.22 & 3.36 \\
\hline
\end{tabular}%
}

\caption{Average word counts and average difference in sentence lengths per record for Paraphrase and Non-Paraphrase datasets.}
\label{tab:avg_word_counts}
\end{table}

\section{Baseline Models}

\subsection{Muril}
MuRIL is a language model built especially for Indian languages and trained entirely on a large volume of Indian language text \citep{khanuja2021muril}. The dataset contains both translated and transliterated document pairings in order to introduce supervised cross-lingual learning during training. 

\subsection{MBERT}
A BERT-based model called Multilingual BERT (mBERT) was trained using text in 104 distinct languages \citep{devlin2019bert}. It is trained with masked language modeling (MLM) and next sentence prediction (NSP) objectives, and it supports a variety of downstream applications, including sentiment analysis.

\subsection{IndicBERT}
Based on the ALBERT architecture \citep{lan2020albertlitebertselfsupervised}, IndicBERT is a language model that was trained on a huge corpus of 12 major Indian languages, including Bengali, English, Gujarati, Hindi, Kannada, Malayalam, Marathi, Oriya, Punjabi, Tamil, Telugu, and Assamese. It employs a combined training technique and makes use of data from the IndicCorp dataset \citep{kunchukuttan2020ai4bharatindicnlpcorpusmonolingualcorpora} in order to better accommodate low-resource languages. There are two versions of the model: IndicBERT (MLM+TLM) and IndicBERT (MLM alone). 
IndicBERT-MLM is trained by masking random tokens in monolingual text and predicting them using context. IndicBERT-TLM uses parallel sentences in different languages, masking tokens and predicting them using both languages.

\subsection{MahaBERT}
A multilingual BERT model called MahaBERT was refined using the L3Cube-MahaCorpus and additional publically accessible Marathi monolingual datasets \citep{joshi2022l3cube1}. 

\section{Result}

The baseline models described above were fine-tuned and evaluated on our dataset, and the results are presented in Table \ref{tab:model_performance}.   
Among the models that were evaluated, MahaBERT was the most accurate model, followed by IndicBERT (MLM + TLM), Muril, IndicBERT (MLM only) and MBERT.

\begin{table}[h!]
\centering
\begin{tabular}{|l|c|c|}
\hline
\textbf{Model} & \textbf{Score} \\ 
\hline

MahaBERT             & 88.7  \\
IndicBERT (MLM+TLM)  & 87.1  \\
Muril                & 86.9  \\
IndicBERT (MLM only) & 85.9  \\
MBERT                & 84.59 \\
\hline
\end{tabular}
\caption{Model Performance Comparison. MLM stands for Masked Language Modeling and TLM stands for Translation Language Modeling.}
\label{tab:model_performance}
\end{table}

\section{Conclusion}
In this paper, we present the MahaParaphrase dataset, comprising of 8,000 labeled pairs of both paraphrase and non-paraphrase sentences. The entire dataset was manually verified by four native Marathi speakers and is further divided into five buckets based on word overlap. These bucketed subsets capture varying degrees of paraphrasing intensity, which can support more nuanced research in this domain.

Furthermore, we evaluate the MahaParaphrase dataset using five baseline models, with MahaBERT achieving the highest performance—an F1 score of 88.7\%.

By providing this low-resource paraphrase dataset, we aim to equip researchers and practitioners with a valuable resource to advance further research in Marathi NLP.

\section{Limitations}

Compared to paraphrase dataset for high-resource languages, this dataset is relatively small (8K pairs). Moreover, the presence of code-mixed sentences introduces minor noise especially when using BERT models trained specifically using Marathi. Additionally, the dataset evaluation was limited to BERT-based models; incorporating LLMs could offer a more comprehensive assessment.

 \section*{Acknowledgement}
This work was carried out under the mentorship of L3Cube, Pune. We would like to express our gratitude towards our mentor, for his continuous support and encouragement. This work is a part of the L3Cube-MahaNLP project \citep{joshi2022l3cube}.

\bibliography{main}

\end{document}